\documentclass[runningheads]{llncs}
\usepackage[T1]{fontenc}
\usepackage{graphicx}
\usepackage{makecell} 
\usepackage{siunitx}
\usepackage{bm}
\usepackage{comment}
\usepackage{xurl}
\usepackage{marvosym}
\usepackage{booktabs}   
\usepackage{multirow}   
\begin{document}

\title{PET/CT Radiogenomic Mutation Prediction in Non-Small Cell Lung Cancer Using Multi-Label Learning}
\titlerunning{PET/CT Raiodgenomic Mutation Prediction in Lung Cancer}

%

%

\author{Mona Furukawa\inst{1}\textsuperscript{(\Letter)}\orcidID{0009-0004-5549-3938} \and Sai Hyne\inst{2} \and Daniel R. McGowan\inst{3,4}\orcidID{0000-0002-6880-5687}\and Bart{\l}omiej W. Papie{\.z}\inst{1}\textsuperscript{(\Letter)}\orcidID{0000-0002-8432-2511} }  
\institute{Big Data Institute, Nuffield Department of Population Health, University of Oxford, Oxford, UK \\
    \email{mona.furukawa@new.ox.ac.uk, bartlomiej.papiez@bdi.ox.ac.uk} 
    \and Department of Radiology, Oxford University Hospitals NHS FT, Churchill Hospital, Oxford, UK
\and
Department of Oncology, University of Oxford, Oxford, UK
\and Department of Medical Physics and Clinical Engineering, Oxford University Hospitals NHS FT, Churchill Hospital, Oxford, UK  
}

\authorrunning{M. Furukawa et al.}

\maketitle              
\begin{abstract}
Lung cancer remains one of the leading causes of cancer-related mortality worldwide. Although targeted therapies have improved outcomes for patients with non-small cell lung cancer (NSCLC), they rely on mutation profiling through tissue biopsy, an invasive procedure with several limitations. This study investigates PET/CT-based radiogenomic prediction of epidermal growth factor receptor (EGFR), tumour protein 53 (TP53), and Kirsten rat sarcoma viral oncogene (KRAS) mutations using deep learning. We further evaluate whether pairwise multi-label learning improves mutation prediction compared with conventional single-gene classification. To the best of our knowledge, this is among the first studies to systematically investigate multi-label learning for PET/CT radiogenomic mutation prediction in NSCLC. Experiments were conducted on a novel UK-based radiogenomics cohort. Joint prediction of KRAS and TP53 improved AUC from 0.58 to 0.64 for KRAS and from 0.69 to 0.71 for TP53. For the EGFR/KRAS pair, only EGFR benefited from joint learning, while no improvement was observed for the EGFR/TP53 pair. 
These findings demonstrate that the effectiveness of multi-label learning depends on the specific combination of gene mutations being modelled, suggesting that mutation-specific modelling strategies may be preferable for PET/CT radiogenomic prediction.

\end{abstract}
\section{Introduction}
Lung cancer is one of the most prevalent malignancies worldwide and remains the leading cause of cancer-related mortality, accounting for approximately 1.8 million deaths annually~\cite{bray2024global}. Despite advances in treatment, patient outcomes remain poor~\cite{lu2019trends}, highlighting the need for more effective precision medicine strategies. Increasingly, treatment decisions are guided by the molecular characteristics of individual tumours rather than a one-size-fits-all approach, reflecting the well-established heterogeneity of cancer and its variable response to therapy~\cite{jeon2025update,krz2018growing}.

An example of such heterogeneity is the presence of epidermal growth factor receptor (EGFR) mutations in a subset of non-small cell lung cancer (NSCLC) patients. Compared with conventional chemotherapy, treatment with EGFR tyrosine kinase inhibitors (TKIs) more than doubles progression-free survival in patients with these mutations, demonstrating substantially improved clinical outcomes~\cite{kara2019egfr}. Co-mutation of TP53 and EGFR has been associated with poorer prognosis in EGFR-mutant patients receiving EGFR TKI therapy, making co-mutation status an important information for treatment planning~\cite{li2025prediction}. 

Tumour mutations are currently identified through tissue biopsy~\cite{shui2021era}, the clinical standard for molecular profiling. However, biopsy has several well-recognized limitations: it captures only a single temporal snapshot of the tumour, samples a limited spatial region that may not capture its full genetic heterogeneity, and can be challenging when lesions are located in difficult-to-access anatomical sites~\cite{rios2017somatic,shui2021era}. Radiogenomics, which aims to infer genomic characteristics from medical imaging, offers a promising non-invasive alternative~\cite{shui2021era}. Unlike biopsy, medical imaging captures the entire tumour~\cite{wang2022predicting} and can be acquired repeatedly over the course of treatment~\cite{rios2017somatic}.

Radiogenomics has evolved from studies investigating associations between semantic imaging features and tumour genetics to quantitative approaches based on hand-crafted radiomics features and machine learning~\cite{Haixian2025}. More recently, deep learning has enabled models to automatically learn of imaging representations from medical images, reducing reliance on manually engineered features~\cite{Haixian2025}. Despite these advances, most NSCLC studies have focused on predicting  the mutation status of individual genes using CT alone~\cite{fuster2026prediction,Haixian2025}. Comparatively few studies have investigated positron emission tomography (PET) and CT-based radiogenomics or explored the simultaneous prediction of multiple genomic alterations~\cite{fuster2026prediction}. PET/CT radiomics were used to predict EGFR/TP53 co-mutation status in lung adenocarcinioma patients~\cite{li2025prediction}, while a multi-task framework was used to predict a panel of 10 molecular biomarkers including programmed death ligand-1 (PD-L1) expression from CT alone~\cite{shao2022radiogenomic}.
Nevertheless, radiogenomic mutation prediction is still predominately formulated as a collection of independent single-gene classification problems. Whether jointly predicting multiple mutations through multi-label learning can improve predictive performance remains largely unexplored.

In this work, we evaluate whether multi-label learning enables more accurate PET/CT-based prediction of EGFR, Kirsten rat sarcoma viral oncogene (KRAS), and TP53 mutations than single-gene classification in NSCLC patients. Our study is motivated by the hypothesis that jointly learning related mutation prediction tasks can improve predictive performance by leveraging shared representations~\cite{mohamed2025deepchest,read2023multi}. 
Jointly learning related mutation prediction tasks is a special case of multi-task learning. We therefore compared independently trained single-gene models with pairwise multi-label models to determine whether shared learning benefits radiogenomics mutation prediction and whether any benefits depend on the combination of genes being predicted. 

Our work makes three main contributions. First, to the best of our knowledge, this is among the first studies to systematically compare single-gene and pairwise multi-label deep learning models for PET/CT-based radiogenomic mutation prediction in NSCLC. Second, we are among the few studies to investigate multi-gene classification using both PET and CT imaging. Third, we evaluate our hypothesis on a novel real-world UK PET/CT radiogenomics cohort, thereby contributing evidence from a patient population that is underrepresented in the radiogenomics literature~\cite{fuster2026prediction}.

\section{Methodology}

\subsection{Dataset} 
The Multimodal PET/CT Imaging dataset for RAdiogenomIc analysis in non-small cell lung cancer (MIRAI dataset)~\cite{mona2026} was used in this study. The dataset contains retrospectively collected, real-world clinical data from a single hospital between November 2014 and October 2024. Following patient selection, the final cohort included 263 patients with a confirmed diagnosis of NSCLC and 293 unique whole-body 18F-FDG PET/CT examinations, with some patients contributing two scans. 

\begin{figure}[t]
\includegraphics[width=\textwidth]{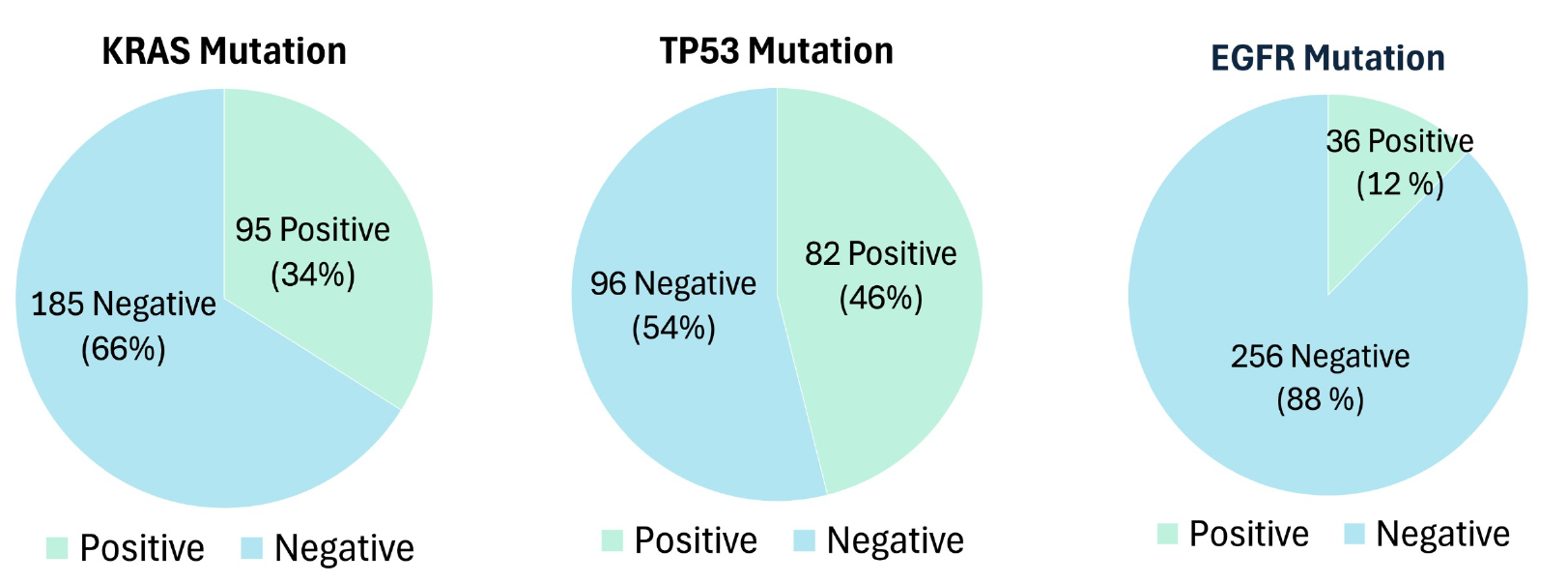}
\caption{Distribution of mutation status for the EGFR, KRAS, and TP53 genes in the study cohort.} \label{gene_dist}
\end{figure}

\begin{table}[t]
\centering
\caption{Patient demographics and clinical characteristics of the study cohort ($N = 263$).}
\label{tab:cohort}
\begin{tabular}{llr}
\toprule
\textbf{Characteristic} & \textbf{Category} & \textbf{N} \\
\midrule
\multirow{3}{*}{Diagnosis}
& Adenocarcinoma & 211 \\
& Non-small cell carcinoma & 43 \\
& Squamous cell carcinoma & 9 \\
\midrule
\multirow{2}{*}{Sex}
& Female & 139 \\
& Male & 124 \\
\midrule
\multirow{3}{*}{Smoking status}
& Current/Former & 133 \\
& Never & 69 \\
& Unknown & 61 \\
\midrule
\multirow{4}{*}{Clinical T stage}
& T1 & 62 \\
& T2 & 50 \\
& T3 & 34 \\
& T4 & 53 \\
\midrule
\multirow{4}{*}{Clinical N stage}
& N0 & 92 \\
& N1 & 32 \\
& N2 & 57 \\
& N3 & 21 \\
\midrule
\multirow{2}{*}{Clinical M stage}
& M0 & 138 \\
& M1 & 61 \\
\bottomrule
\end{tabular}
\end{table}

For each PET/CT, pathologically confirmed mutation status was available, and genetic testing was performed within a time window of 6-month before or after the corresponding PET/CT acquisition. 
A $\pm$ 6-month interval between PET/CT acquisition and genetic testing was selected to maximise the number of imaging-genomic pairs while maintaining clinical relevance.
Mutation data were available for the EGFR, KRAS, and TP53 genes; however, not all patients were tested for all three genes. A summary of the cohort demographics, clinical characteristics, and mutation status is provided in Table~\ref{tab:cohort} and Fig.~\ref{gene_dist}. 

PET/CT were acquired using either GE HealthCare's Discovery 710 or Discovery 690 scanner. Furthermore, patients received an intravenous 18F-FDG injection of 4 MBq/kg, followed by 90-minute uptake period. The PET volumes were reconstructed using the GE HealthCare’s Q.Clear Bayesian penalised likelihood reconstruction algorithm~\cite{teoh2015phantom}. CT images were acquired as part of the clinical PET/CT protocol and used for PET attenuation correction and anatomical localisation.

\subsection{Tumour annotations} 
\begin{figure}[t]
\includegraphics[width=\textwidth]{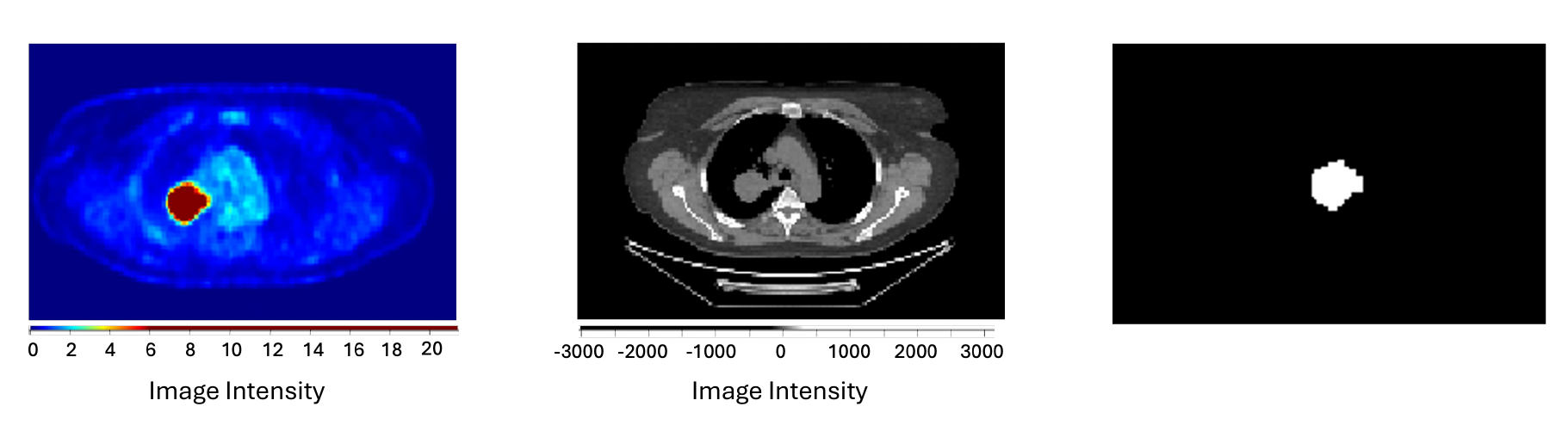}
\caption{Example axial view of PET (SUV) and CT (HU) images with the corresponding manual tumour segmentation from the study dataset.} \label{fig1}
\end{figure}

A total of 130 NSCLC PET/CT volumes were manually segmented. Of these, 108 had corresponding gene mutation labels available and were included in the final gene mutation prediction dataset. 
Tumours were manually delineated slice-by-slice on PET volumes using ITK-SNAP~\cite{py06nimg}, with the corresponding CT volumes and radiology reports used to guide the segmentation process. All manual segmentations were subsequently reviewed and verified by an experienced radiologist. An example of a PET/CT and its corresponding manual tumour segmentation is displayed in Fig.~\ref{fig1}.

The remaining PET/CT volumes in the gene mutation prediction dataset were automatically segmented using a trained nnU-Net segmentation model~\cite{isensee2021nnu}. The model was pretrained on 168 NSCLC PET/CT volumes from the MICCAI autoPET challenge 2022~\cite{gatidis2022whole} and fine-tuned using 110 manually segmented cases, while the remaining 20 manually segmented cases were reserved for evaluation. 
Predicted segmentations were post-processed by removing tumours located outside the lungs using the lung cavity mask predicted by the TotalSegmentator~\cite{2023totalsegmentator}.  

\subsection{Image Preprocessing}
PET/CT volumes and their corresponding segmentation masks were resampled to an isotropic voxel spacing of 3.27~$mm$ using linear interpolation for PET/CT and nearest-neighbour interpolation for the segmentation masks. Lymph node and metastatic lesion annotations were removed, retaining only the primary tumour mask for multi-label mutation prediction.  

\subsection{Multi-gene and single gene classification model} 
A ResNet-18~\cite{he2016deep} pretrained on ImageNet-1K was used as the backbone for all classification models. The final fully connected layer was replaced with either a single output for single-gene classification or two outputs for two-gene classification. Rather than freezing the pretrained backbone, all network parameters were updated during training. For each gene pair, only patients with mutation status available for both genes were included in the analysis. Patients were divided into training, validation, and test sets using a stratified 70:15:15 split based on the target labels. 
Label combinations occurring fewer than six times across the entire cohort were excluded because they were unlikely to be consistently represented across the subsets. Furthermore, repeated PET/CT examinations from the same patient were treated as independent samples and randomly assigned to the training, validation, and test sets. The resulting cohort sizes for each gene pair are summarised in Table~\ref{tab:split}. To enable a fair comparison, the corresponding single-gene models were trained and evaluated using the same patient split as the associated multi-gene cohort. 

For each patient, the axial PET and CT slices corresponding to the largest cross-sectional area of the primary tumour were extracted together with the associated tumour segmentation mask. A bounding box was generated from the segmentation mask and used to crop the PET and CT slices, which were subsequently resized to $224 \times 224$ pixels.
The PET, CT, and segmentation mask were concatenated along the channel dimension to form the network input. 
CT slices were intensity-clipped to the 0.5th and 99.5th percentiles of the foreground voxel intensities in the training set followed by Z-score normalisation using the mean and standard deviation computed from the foreground voxel intensities of the training set. PET slices were Z-score normalised using the global mean and standard deviation of the training data. During training, random horizontal flips, affine transformations, and Gaussian noise were applied as data augmentation.

The models were trained using the binary cross-entropy loss with logits, which combines a sigmoid activation with the binary cross-entropy loss. To mitigate class imbalance, a positive weight was assigned to the loss and was calculated separately for each gene using the training data \cite{BCElogit}. Models were trained for a maximum of 30 epochs with a batch size of 5, and early stopping with a patience of 10 epochs was applied based on the validation loss. Each model was trained using five different random seeds. Performance was evaluated on the test set using the area under the curve (AUC), precision, recall and F1-score. 

\begin{table}[t]
\centering
\footnotesize
\caption{Joint mutation status for each gene pair.  Each cell shows the number of PET/CT examinations and the corresponding percentage of the cohort. EGFR and KRAS are mutually exclusive in this cohort (no patient carried both mutations). $N$ denotes the number of PET/CT examinations with valid tumour segmentations and mutation labels available for both genes after excluding rare label combinations (<6 cases) to enable stratified data splitting.}
\label{tab:split}
\setlength{\tabcolsep}{10pt}
\renewcommand{\arraystretch}{1.2}
\begin{tabular}{l l c c}
\toprule
Gene pair & & \multicolumn{2}{c}{} \\
\midrule
\multirow{3}{*}{\makecell[l]{\\\\ \textbf{EGFR/KRAS}\\\footnotesize $N=280$}}
   &         & KRAS+ & KRAS$-$ \\
\cmidrule(lr){3-4}
 & EGFR+   & \makecell{\textbf{0}\\\footnotesize(0.0\%)}  & \makecell{26\\\footnotesize(9.3\%)}  \\
 & EGFR$-$ & \makecell{95\\\footnotesize(33.9\%)} & \makecell{159\\\footnotesize(56.8\%)} \\
\midrule
\multirow{3}{*}{\makecell[l]{\\\\ \textbf{KRAS/TP53}\\\footnotesize $N=178$}}
   &         & KRAS+ & KRAS$-$ \\
\cmidrule(lr){3-4}
 & TP53+   & \makecell{31\\\footnotesize(17.4\%)} & \makecell{51\\\footnotesize(28.7\%)} \\
 & TP53$-$ & \makecell{33\\\footnotesize(18.5\%)} & \makecell{63\\\footnotesize(35.4\%)} \\
\midrule
\multirow{3}{*}{\makecell[l]{\\\\ \textbf{EGFR/TP53}\\\footnotesize $N=173$}}
   &         & TP53+ & TP53$-$ \\
\cmidrule(lr){3-4}
 & EGFR+   & \makecell{--\textsuperscript{a}\\\footnotesize{}} & \makecell{10\\\footnotesize(5.8\%)} \\
 & EGFR$-$ & \makecell{77\\\footnotesize(44.5\%)} & \makecell{86\\\footnotesize(49.7\%)} \\
\midrule
\multicolumn{4}{l}{\footnotesize \textsuperscript{a}\,EGFR+/TP53+ excluded: below the minimum count for stratified splitting.}\\
\bottomrule
\end{tabular}
\end{table}

\section{Results and Discussion}

\begin{table}[t]
\centering
\caption{Segmentation performance of the pretrained nnU-Net model on the test set.}
\setlength{\tabcolsep}{12pt}
\begin{tabular}{ccc}
\toprule
\textbf{Dice} & \textbf{IoU} & \textbf{HD95 (mm)} \\
\midrule
$0.78 \pm 0.12$ & $0.65 \pm 0.15$ & $9.19 \pm 15.10$ \\
\bottomrule
\end{tabular}
\label{seg_metric}
\end{table}

The pretrained nnU-Net model was tested on the remaining 20 manually segmented tumours and the segmentation performance is shown in Table~\ref{seg_metric}. 
The model achieved a mean Dice coefficient of 0.78, which is comparable to the performance of the seven highest-ranked teams in the autoPET 2022 challenge, whose mean Dice coefficients ranged between 0.74 and 0.79~\cite{gatidis2024results}. These results indicate that the model provides sufficient segmentation accuracy for the automatic annotation of the remaining PET/CT examinations in the mutation prediction dataset.

\begin{table}[t]
\centering
\footnotesize
\caption{Comparison of single-gene and pairwise multi-label mutation prediction on PET/CT. Within each gene pair, the single-gene and multi-label models were trained and evaluated using the same stratified 70:15:15 patient split. Values are reported as mean ± standard deviation over five random seeds. Bold indicates the better-performing model for each gene within a pair.}
\label{tab:results}
\setlength{\tabcolsep}{1.5pt}
\renewcommand{\arraystretch}{1}

\begin{tabular}{l l l S S S S}
\toprule
\shortstack{Gene\\pair} & {\bfseries Model} & {\bfseries Gene} &
{\bfseries AUC} &
{\bfseries Precision} &
{\bfseries Recall} &
{\bfseries F1} \\
\midrule

\multirow{4}{*}{\shortstack{EGFR\\\&\\KRAS}}
 & \multirow{2}{*}{Joint} & EGFR &
 {$\bm{0.746 \pm 0.093}$} &
 {$\bm{0.186 \pm 0.162}$} &
 {$\bm{0.550 \pm 0.326}$} &
 {$\bm{0.267 \pm 0.205}$} \\
 &                       & KRAS &
 {$0.557 \pm 0.114$} &
 {$\bm{0.354 \pm 0.065}$} &
 {$0.614 \pm 0.199$} &
 {$\bm{0.446 \pm 0.098}$} \\
\cmidrule(lr){2-7}
 & \multirow{2}{*}{Single} & EGFR & {$0.718\pm0.057$} & {$0.154\pm0.092$} & {$0.400\pm0.224$} & {$0.221\pm0.128$} \\
 &                         & KRAS & {$\bm{0.632\pm0.104}$} & {$0.293\pm0.181$} & {$\bm{0.629\pm0.456}$} & {$0.390\pm0.255$} \\
\midrule
\multirow{4}{*}{\shortstack{KRAS\\\&\\TP53}}
 & \multirow{2}{*}{Joint}  & KRAS & {$\bm{0.641\pm0.050}$} & {$\bm{0.479\pm0.058}$} & {$0.540\pm0.182$} & {$\bm{0.499\pm0.108}$} \\
 &                         & TP53 & {$\bm{0.712\pm0.038}$} & {$\bm{0.614\pm0.057}$} & {$\bm{0.600\pm0.227}$} & {$\bm{0.584\pm0.095}$} \\
\cmidrule(lr){2-7}
 & \multirow{2}{*}{Single} & KRAS & {$0.575\pm0.068$} & {$0.394\pm0.070$} & {$\bm{0.660\pm0.167}$} & {$0.489\pm0.091$} \\
 &                         & TP53 & {$0.689\pm0.100$} & {$0.610\pm0.164$} & {$0.523\pm0.227$} & {$0.542\pm0.163$} \\
\midrule

\multirow{4}{*}{\shortstack{EGFR\\\&\\TP53}}
 & \multirow{2}{*}{Joint}  & EGFR & {$0.783\pm0.357$} & {$\bm{0.141\pm0.071}$} & {$\bm{0.900\pm0.224}$} & {$\bm{0.241\pm0.112}$} \\
 &                         & TP53 & {$0.491\pm0.048$} & {$\bm{0.435\pm0.016}$} & {$\bm{0.582\pm0.209}$} & {$\bm{0.485\pm0.069}$} \\
\cmidrule(lr){2-7}
 & \multirow{2}{*}{Single} & EGFR & {$\bm{0.808\pm0.226}$} & {$0.127\pm0.060$} & {$0.800\pm0.274$} & {$0.218\pm0.099$} \\
 &                         & TP53 & {$\bm{0.514\pm0.107}$} & {$0.354\pm0.202$} & {$0.564\pm0.377$} & {$0.421\pm0.239$} \\
\bottomrule

\end{tabular}
\end{table}
Table~\ref{tab:results} summarises the performance of the  single-gene and multi-label mutation classification. For the KRAS and TP53 pair, joint learning shows improved performance for both tasks, with the AUC increasing from 0.575 to 0.641 for KRAS and from 0.689 to 0.712 for TP53 compared with the corresponding single-gene models. This finding suggests that the two tasks benefit from shared representation learning, resulting in positive transfer between KRAS and TP53 prediction tasks~\cite{mohamed2025deepchest}.  
For the EGFR/KRAS pair, joint learning improved EGFR prediction but reduced KRAS performance, suggesting that the two tasks do not benefit equally from shared representation learning and that negative transfer may occur for KRAS~\cite{mohamed2025deepchest}. This is somewhat counterintuitive given the class imbalance: where only 9.3\% of the total sample is KRAS-negative/EGFR-positive. Here, we would normally expect the dominant class (KRAS-positive/ EGFR-negative or KRAS-negative/EGFR-negative) to dominate learning, yet we observe the opposite. One possible explanation is that the larger positive class weight assigned to EGFR influences the optimisation process more, resulting in task imbalance during training~\cite{mohamed2025deepchest,BCElogit}. Alternatively, imaging features associated with EGFR mutation may be more discriminative than those associated with KRAS mutation, causing the shared representation to favour EGFR prediction.
This interpretation is consistent with previous radiogenomics studies, which have reported more reproducible imaging correlates for EGFR than for KRAS mutations ~\cite{pinheiro2020,rizzo2019genomics,yip2017associations}. This difference is also reflected in the single-gene models, where KRAS prediction achieved lower performance than EGFR prediction. For the EGFR/KRAS pair, these results suggest that a hybrid strategy may be preferable, using joint learning to improve EGFR prediction while modelling KRAS independently.

For the TP53/EGFR pair, the joint learning degraded performance for both tasks, with the corresponding single-gene models achieving higher AUCs. This suggests that these tasks may share limited complementary information and therefore derive little benefit from shared representation learning~\cite{mohamed2025deepchest}. Future work could investigate if gradient conflict occurs in this setting, whereby gradients from the two tasks compete during optimisation. Another contributing factor may be the severe class imbalance in the EGFR task with EGFR-positive samples accounting for 5.8\% of the cohort. Interestingly, although the joint model achieved lower AUC, it produced slightly higher precision, recall, and F1-scores than the corresponding single-gene models. This suggests that the shared representation may still provide limited benefits for classification. However, these findings should be interpreted cautiously, as no EGFR-positive/TP53-positive cases remained in the modelling dataset after preprocessing. Larger cohorts with sufficient dual-mutant cases are needed to validate these observations in future studies. Overall, these findings suggest that multi-label learning should not be adopted uniformly for radiogenomic mutation prediction. Instead, mutation-specific modelling strategies may be preferable, with joint learning applied only when it provides measurable performance gains.

\section{Conclusion}
In conclusion, this study demonstrates that the effectiveness of multi-label learning for PET/CT-based radiogenomic mutation prediction depends on the specific combination of gene mutations being modelled. Specifically, joint learning improved the prediction of KRAS and TP53 mutations but did not improve performance for the TP53/EGFR pair. These findings suggest that multi-label learning should not be applied uniformly across all radiogenomic prediction tasks. Instead, mutation-specific modelling strategies may be preferable, with joint learning adopted only when it provides measurable performance gains.

This study has several limitations. Firstly, the dataset exhibits class imbalance, which may have affected model performance. However, this distribution reflects the prevalence of these mutations in a real-world UK NSCLC population, enhancing the clinical relevance of the study. Secondly, the study was conducted using data from a single centre without external validation, which may limit the generalisability of the findings~\cite{zhang2021}. Thirdly, tissue biopsy samples were obtained from a single region of each tumour, which may not adequately capture the full spatial heterogeneity of the tumour~\cite{shui2021era}. Fourthly, this work was limited to single 2D image analysis and did not explore the use of 3D models. Future work will investigate strategies to better address class imbalance and incorporate approaches that effectively represent tumour heterogeneity, validate the proposed approach on external multi-centre cohorts, and explore the use of 3D models.

%

\begin{credits}
\subsubsection{\ackname} Supported by EPSRC(EP/\allowbreak S024093/\allowbreak 1), SABS R3 CDT University of Oxford, and GE HealthCare. BWP acknowledges MRC funding (MR/\allowbreak Y008421/\allowbreak 1). Retrospective anonymised NHS data were accessed via the Thames Valley and Surrey Secure Data Environment, part of the NHS Research Secure Data Environment Network, with Health Research authority approval (24/HRA/1337).

\subsubsection{\discintname}
None.
\end{credits}
%
%
%
\bibliographystyle{splncs04}
\bibliography{mybibliography}

\end{document}